%% file: CADI-AI.tex
\documentclass{article}

\usepackage{microtype}
\usepackage{graphicx}
\usepackage{subfigure}
\usepackage{booktabs} 

\usepackage{hyperref}
\usepackage{float}
\usepackage{xcolor}

\newcommand{\anon}[1]{#1}

\usepackage[accepted]{icml2023}

\usepackage{amsmath}
\usepackage{amssymb}
\usepackage{mathtools}
\usepackage{amsthm}
\usepackage{comment}
\usepackage{multicol}
\usepackage{blindtext}
\usepackage{wrapfig}
\usepackage[font=normalfont,labelfont=bf]{caption}

\usepackage[capitalize,noabbrev]{cleveref}

\PassOptionsToPackage{unicode}{hyperref}
\PassOptionsToPackage{hyphens}{url}
\usepackage{amsmath,amssymb}

\theoremstyle{plain}

\theoremstyle{definition}

\theoremstyle{remark}

\usepackage[textsize=tiny]{todonotes}

\icmltitlerunning{Cashew Disease Identification With Artificial Intelligence (CADI AI)}

\begin{document}

\twocolumn[

\icmltitle{Localized Data Work as a Precondition for Data-Centric ML: \\
A Case Study of Full Lifecycle Crop Disease Identification in Ghana}



\begin{icmlauthorlist}
\icmlauthor{Darlington Akogo}{kara}
\icmlauthor{Issah Samori}{kara}
\icmlauthor{Cyril Akafia}{kara}
\icmlauthor{Harriet Fiagbor}{kara}
\icmlauthor{Andrews Kangah}{kara}
\icmlauthor{Donald Kwame Asiedu}{kara}
\icmlauthor{Kwabena Fuachie}{kara}
\icmlauthor{Luis Oala}{dotp}
\end{icmlauthorlist}

\icmlaffiliation{kara}{KaraAgro AI, Accra, Ghana}
\icmlaffiliation{dotp}{Dotphoton AG, Zug, Switzerland}

\icmlcorrespondingauthor{Harriet Fiagbor}{harrietfiagbor@gmail.com}
\icmlcorrespondingauthor{Cyril Akafia}{kwakucyril@gmail.com}

\icmlkeywords{Machine Learning, ICML}
\vskip 0.3in
]



\printAffiliationsAndNotice{\icmlEqualContribution} 

\begin{abstract}
The Ghana Cashew Disease Identification with Artificial Intelligence (CADI AI) project demonstrates the importance of sound data work as a precondition for the delivery of useful, localized data-centric solutions for public good tasks such as agricultural productivity and food security. Drone-collected data and machine learning are utilized to determine crop stressors. Data, model and the final app are developed jointly and made available to local farmers via a desktop application.
\end{abstract}
\vspace{-.4cm}



Cashew is a significant cash crop in Ghana \cite{rabany2015african}, with small and medium farmers relying on it for income. Cashew cultivation is concentrated in specific regions of Ghana. However, farmers face challenges including insect, plant disease and abiotic stress factors that reduce their yields \cite{cashewic17:online,jayaprakash2023cashew,mensah2023ccmt,timothy2021detection}. To address these issues, the Cashew Disease Identification With Artificial Intelligence (CADI AI) project was launched to provide a data-centric solution. The project encompasses three stages. Comprehensive data work encompassed stakeholder consultation, data collection, data annotation and labelling. The collected drone data is open for researchers and data scientists to develop innovative machine learning applications to improve food security. Model work involved the training of an object detection model to diagnose and detect stress factors in cashew crop images. Finally, the model was integrated into a desktop application for farmers, allowing them to input their own data and receive diagnoses. The application also displays the precise location of the image, enabling farmers to identify affected areas on their farms.

\begin{figure}[ht]
    \centering
    \includegraphics[width=\linewidth]{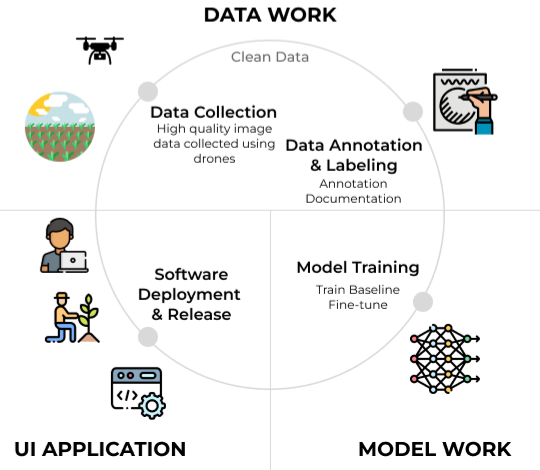}
    \caption{A visual summary of the application lifecycle: data work (data collection with farmers, data annotation and labelling), model work (model training and fine-tuning), and UI application (software deployment and release to farmers).}
    \label{fig:cadiflow}
\end{figure}

\section{Data work}

The data was collected from cashew farms in the Bono Region of Ghana, necessitating two separate trips to the farms to accommodate seasonal variations and diversity of data. In total, the data collection process spanned six days. 
The dataset is diverse in maturity stages, camera angles, time of capture, and various types of stress morphology. All images were captured with the P4 Multi-spectral drone \cite{dji} at image resolution of 1600 x 1300 pixels. The images comprise close up shots and shots from distance of the cashew plant abnormalities. The total number of images collected is 4,736. Full details and datasheet in the appendices. Further improvements to the dataset could be made by capturing across more regions during blooming cycles or varying devices for robustness  testing \cite{oala2023data}.

\subsection{Data Annotation/Labelling}

\begin{figure}[t]
    \centering
    \includegraphics[width=8cm]{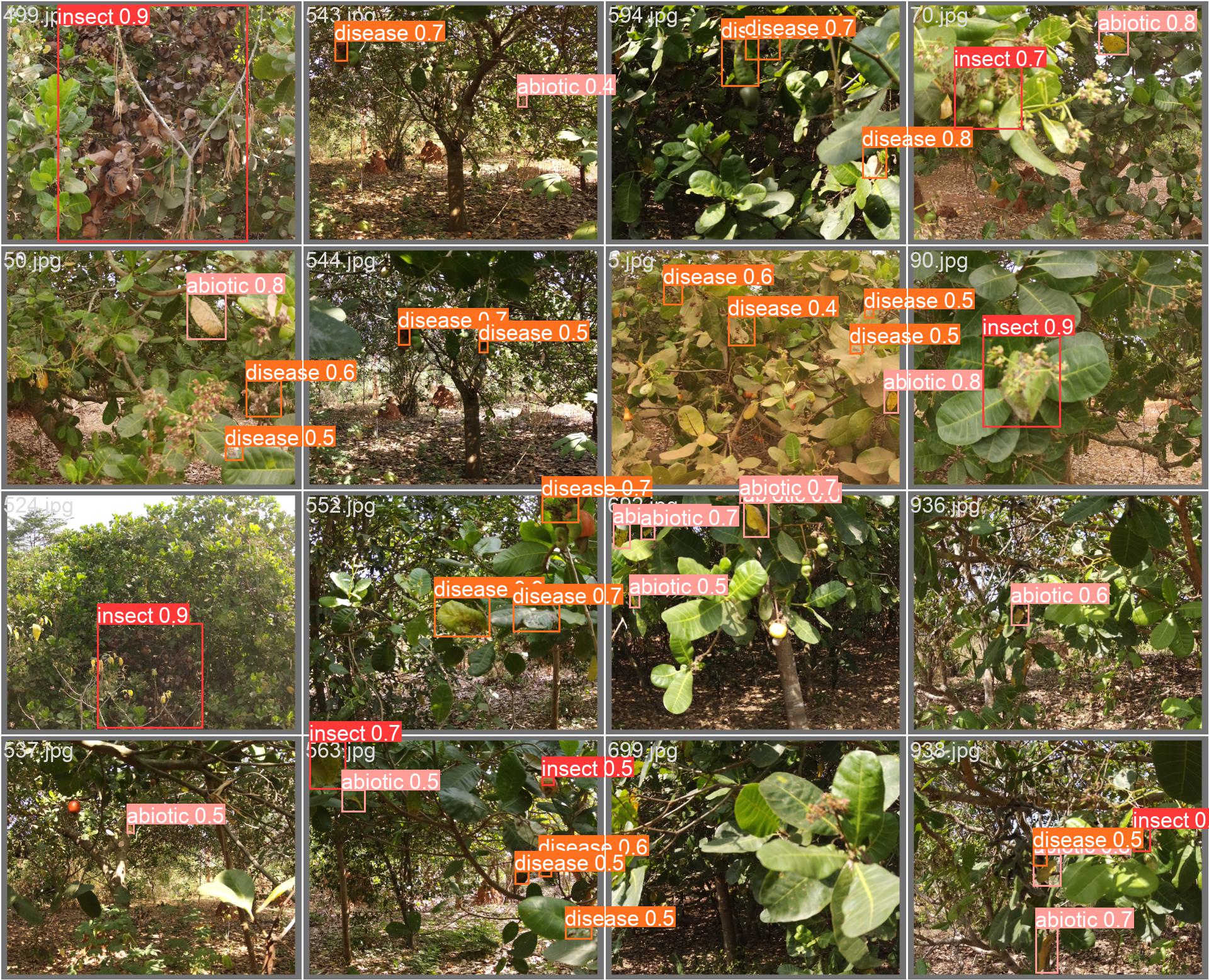}
    \includegraphics[width=8cm]{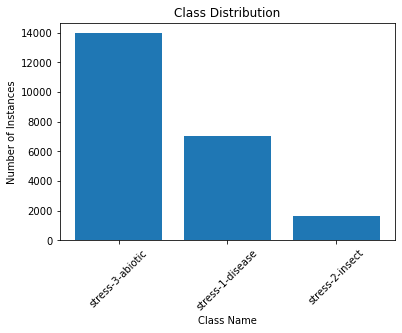}
    \caption{Top: Sample instances from the annotated dataset. For a higher resolution sample see the appendices. Bottom: Distribution of labels in the annotated data.}
    \label{fig:annotated-dist}
    \vspace{-.8cm}
\end{figure}

The data was annotated by the project team with labelling tools makesense \cite{make-sense} and roboflow \cite{roboflow}.
Refer to appendix A.1 for annotation guidelines developed by an agricultural scientist with expertise in crop health and disease management from a local Ghanaian university.
Each stress instance is associated with a class label based on the status of the crop. The labels are \textbf{“insect”}, \textbf{“disease”} and \textbf{“abiotic”} respectively as depicted in \Cref{fig:annotated-dist}.
The data was split into train, validation and test sets i.e., 3788, 710 and 238 images respectively. During the training, it was found that the dataset is significantly skewed towards the abiotic class. 

\section{Model work}

We utilized the YOLO v5X \cite{yolov5} architecture, known for its strong performance on object detection benchmarks, as the foundation for this study. The experiments were conducted on the high-performance DFKI GIZ cluster.
The dataset has a significant skew towards the abiotic class (see \Cref{fig:annotated-dist}), and measures were taken to balance the data by augmenting other classes. However, preserving the skewness was important to reflect the higher occurrence of abiotic factors on farms. The best model achieved a mean average precision (mAP) of 0.648. See \Cref{tab:yolo} and \Cref{fig:conf-mat} in the Appendices for detailed experimental evaluations and baselines.
%
%
The model has a few limitations that affect its performance in distinguishing between the disease class and the abiotic class. The primary challenge lies in the similarity between these two classes within a typical farm setting. The model may encounter difficulties in accurately differentiating between them due to their overlapping characteristics. This limitation is an inherent challenge in the dataset and can impact the model's accuracy when classifying these instances.
However, it is worth noting that the model exhibits strong performance when it comes to the insect class. This is attributed to the distinct characteristics of insect class, which make them easier to identify and classify accurately.

\section{Closing the loop: UI application}

\begin{figure}
    \centering
    \includegraphics[width=8cm]{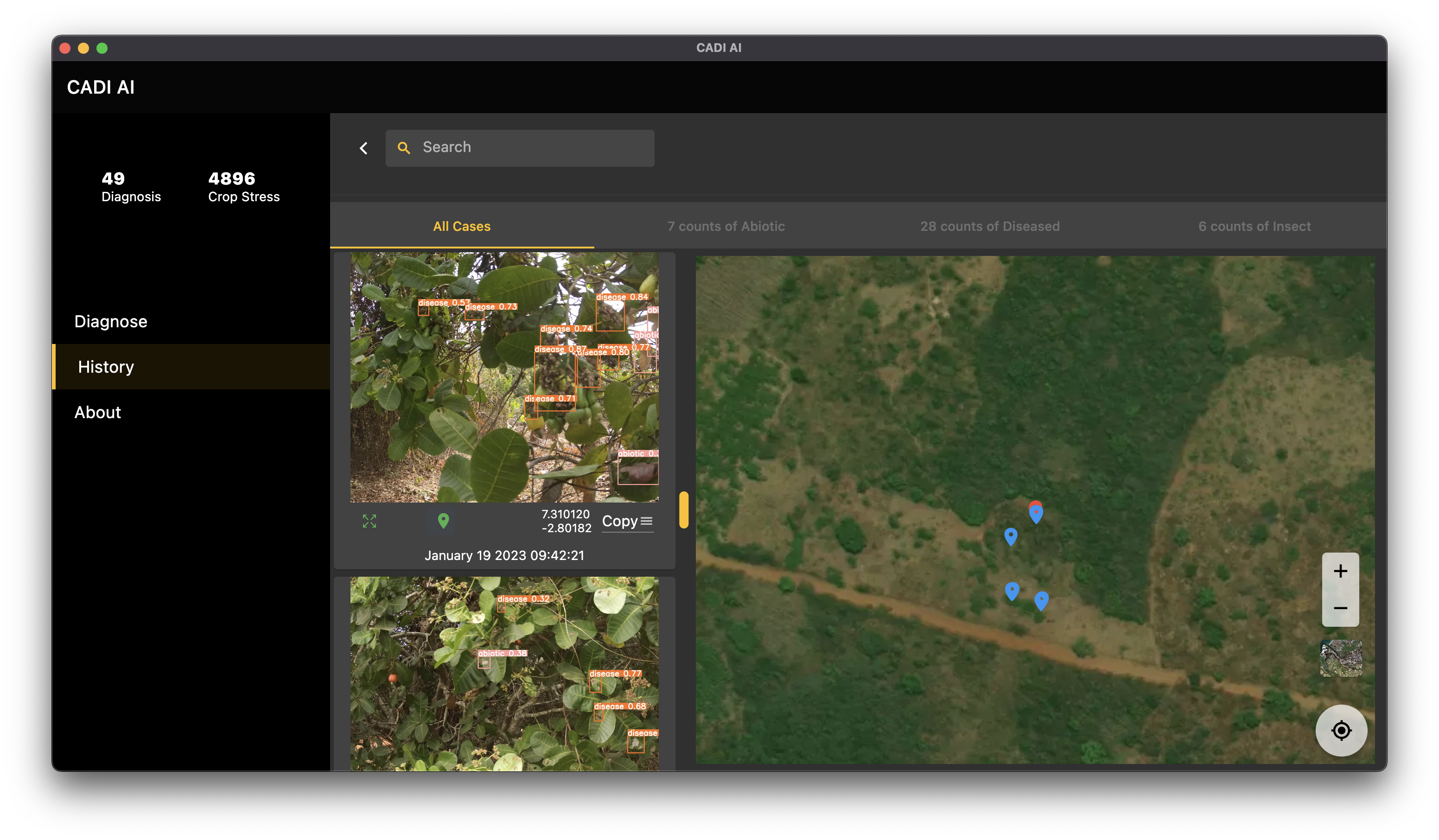}
    \caption{Screenshot of the final UI application. For more details see appendices.}
    \label{fig:enter-label}

\end{figure}
A software application built with Flutter \cite{google_2019} was infused with the CNN model trained on the collected data, allowing farmers to use the model on new data for crop disease management. 
%
Our objective is to make the CADI AI project's data, model, and software widely accessible to maximize their impact within the agriculture community. The data is available through prominent platforms such as Kaggle  and Hugging Face dataset hub. These platforms provide user-friendly interfaces, allowing researchers, developers, and enthusiasts to access the data for their specific needs. Furthermore, the model itself can be accessed through the Hugging Face platform, enabling users to leverage its capabilities in their own ML applications. 
A summary of all resources is in the appendices.

\bibliography{CADI-AI}
\bibliographystyle{icml2023}

\newpage
\appendix
\onecolumn
\section{Data Annotation}

\subsection{Guidelines for Annotation}
Below are some guidelines for labeling or annotating the data set by team.
\begin{itemize}
    \item Provide enough space to capture the affected area without cutting any part off, but avoid introducing too much space.
    \item If possible, zoom into the affected area to ensure accurate annotation.
    \item Avoid including too much gap between two affected areas. Treat a wide gap as another annotation.
    \item To ensure completeness, try to annotate every possible instance as long as it is visible.
    \item Even for images that appear to have no affected areas, zoom in and carefully examine the image to ensure nothing is missed. It is better to have an annotation than to remove the image altogether. However, if truly nothing exists, discard the image from the database.
    \item Avoid using images with human faces whenever possible.
    \item If an image with a human face is necessary, but the face is not centered, consider cropping it out and replacing the original image with the cropped version
\end{itemize}

\subsection{Deciding the Labels}
\begin{itemize}
    \item \textbf{Insect/ pest stress factors} represent the damage to crops by insects or pests
    \item \textbf{Diseased factors} represent attacks on crops by microorganisms.
    \item \textbf{Abiotic stress factors} represent stress factors caused by non-living factors, e.g. environmental factors like weather or soil conditions or the lack of mineral nutrients to the crop.
\end{itemize}

The decision to use the labels "abiotic", "disease", and "insect" for our object detection task was recommended by an agricultural scientist with  expertise in crop health and disease management, \anon{Dr. Torkpor Stephen} from \anon{University of Ghana}.

\section{Evaluation and Results}
\begin{table}[!ht]
    \centering
    \caption{Table showing Experimental Procedures using YOLO architecture}
    \scalebox{0.6}{
    \begin{tabular}{lllllll}
    \toprule
        \textbf{Training Data} & \textbf{Pretrained Model Used} & \textbf{Epoch} & \textbf{Image size} & \textbf{Batch Size} & \textbf{Comments} & \textbf{mAP} \\ \midrule
        \textbf{Original and background images} & yolov8x & 30 & 640 & 16 & Little overfitting & 0.6 \\
        ~ & yolov8x & 30 & 640 & 16 & More background images were included. This lead to overfiting. & 0.56 \\
        ~ & yolo8m & 30 & 640 & 16 & Tried the above experiment with a smaller architecture & 0.57 \\ 
        ~ & ~ & ~ & ~ & ~ & ~ & ~ \\
        \textbf{Only original images} & yolov8m & 50 & 640 & 16 & Early overfitting & 0.59 \\ 
        ~ & yolov8n & 50 & 640 & 16 & Did not converge & 0.53 \\ 
        ~ & yolov8l & 30 & 640 & 16 & Overfitting & 0.58 \\
        ~ & yolov8x & 30 & 640 & 16 & Overfitting & 0.61 \\ 
        ~ & yolov8n & 60 & 640 & 16 & Did not converge & 0.55 \\ 
        ~ & yolov8n & 100 & 640 & 16 & Overfitting & 0.57 \\ 
        ~ & yolov8m & 50 & 960 & 16 & Overfitting & 0.59 \\ 
        ~ & yolov8n & 80 & 1280 & 16 & Overfitting & 0.58 \\ 
        ~ & ~ & ~ & ~ & ~ & ~ & ~ \\ 
        \textbf{Original, background and} & yolov5x & 30 & 640 & 16 & Converged & 0.6 \\ 
        \textbf{augmented images}  & yolov5t & 45 & 640 & 16 & Overfitting & 0.58 \\ 
        ~ & yolov5m6 & 50 & 640 & 16 & Overfitting & 0.572 \\ 
        ~ & yolov5x & 35 & 640 & 32 & Increased batch size lead to increased mAP & 0.615 \\
        ~ & yolov5x & 35 & 640 & 64 & Overfitting & 0.601 \\ 
        ~ & yolov5x & 35 & 640 & 48 & Little Overfitting & 0.603 \\ 
        ~ & \textbf{yolov5x} & \textbf{25} & \textbf{640} & \textbf{56} & \textbf{Batch size lead to a significant increase in performance. The model converged.} & \textbf{0.648} \\ 
        ~ & yolov5x & 40 & 640 & 56 & Overfitting & 0.606 \\
        ~ & yolov5x & 40 & 640 & 72 & Overfitting & 0.616 \\ 
        ~ & yolov5x & 40 & 640 & 72 & Overfitting& 0.616 \\ 
        ~ & yolov5x & 35 & 640 & 48 & Overfitting & 0.58 \\ 
        ~ & yolov5x & 35 & 640 & 48 & Overfitting & 0.57 \\ 
        ~ & yolov8x & 40 & 640 & 32 & Overfitting & 0.55 \\ 
        ~ & yolov8n & 30 & 640 & 32 & Overfitting & 0.55 \\ 
        ~ & yolov8n & 30 & 640 & 32 & Overfitting & 0.48 \\ 
        ~ & yolov8n & 50 & 640 & 32 & Overfitting & 0.53 \\ 
        \bottomrule
    \end{tabular}
    }
    \label{tab:yolo}
\end{table}

\begin{figure}[H]
    \centering
    \includegraphics[width=.33\textwidth]{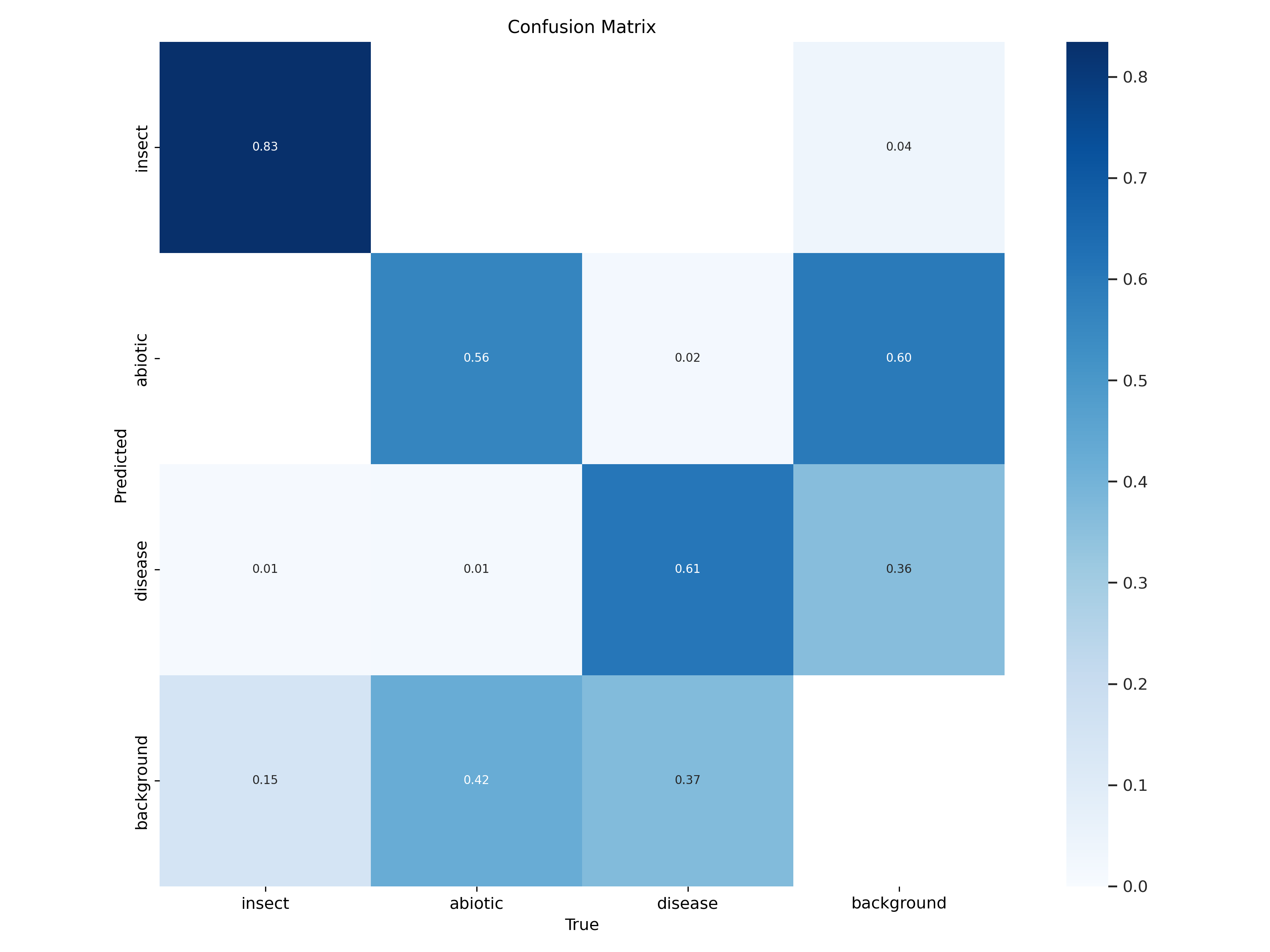}
    \includegraphics[width=.33\textwidth]{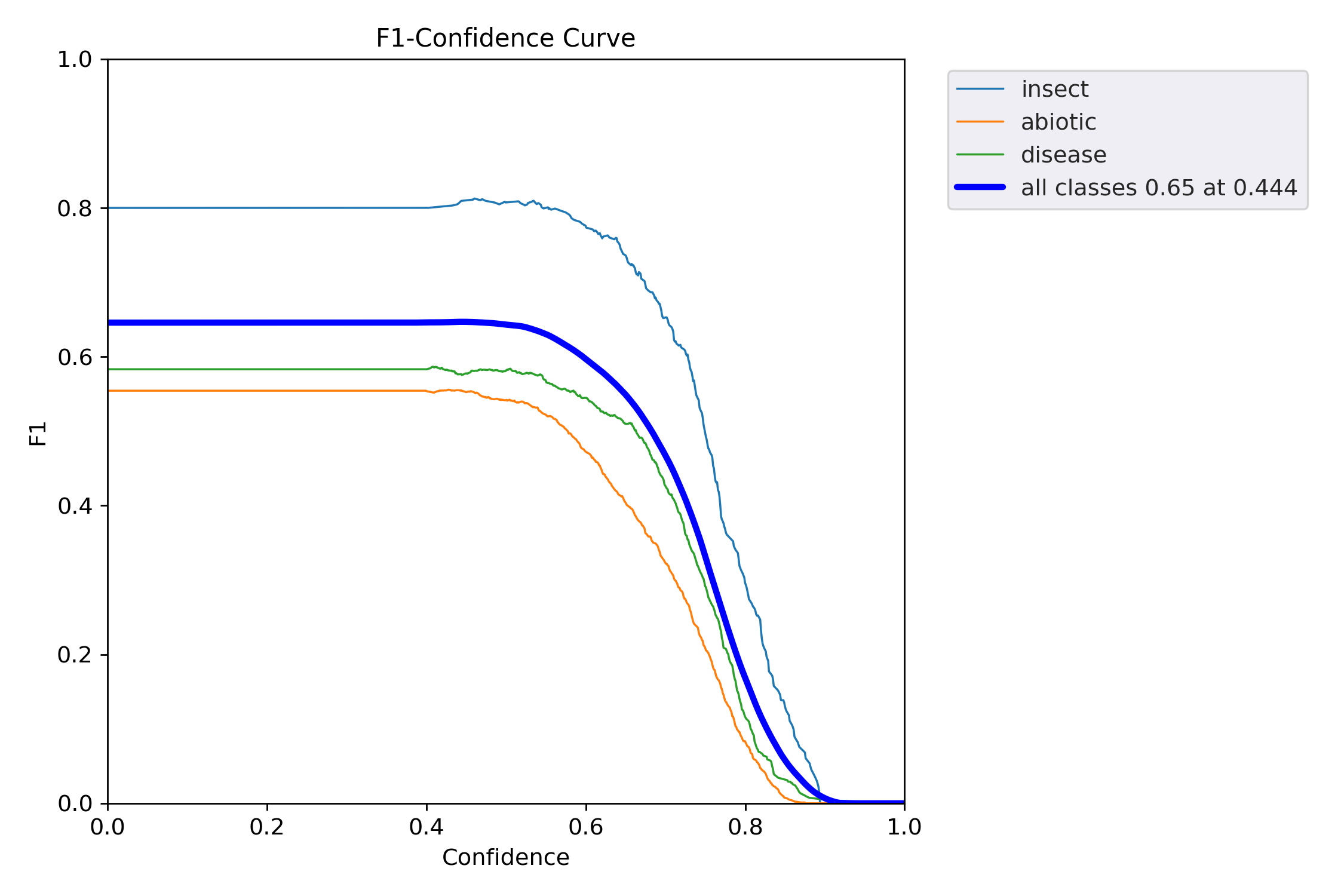}
    \includegraphics[width=.33\textwidth]{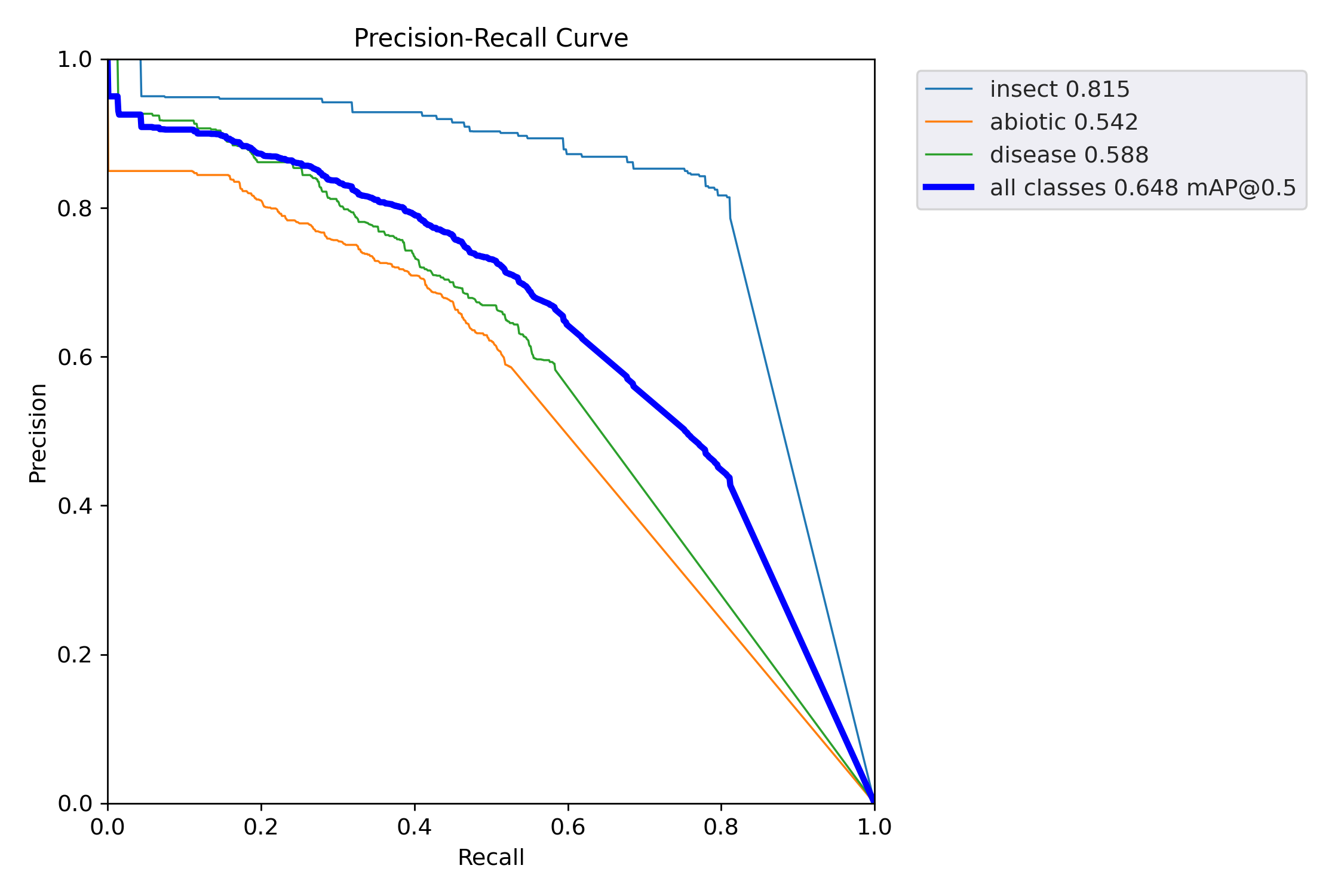}
    \caption{Left: Confusion Matrix of Evaluation results. Center: F1-Curve of Evaluation results. Right: PR-Curve of Evaluation results}
    \label{fig:conf-mat}
\end{figure}

\begin{figure}[H]
    \centering
    \includegraphics[width=\textwidth]{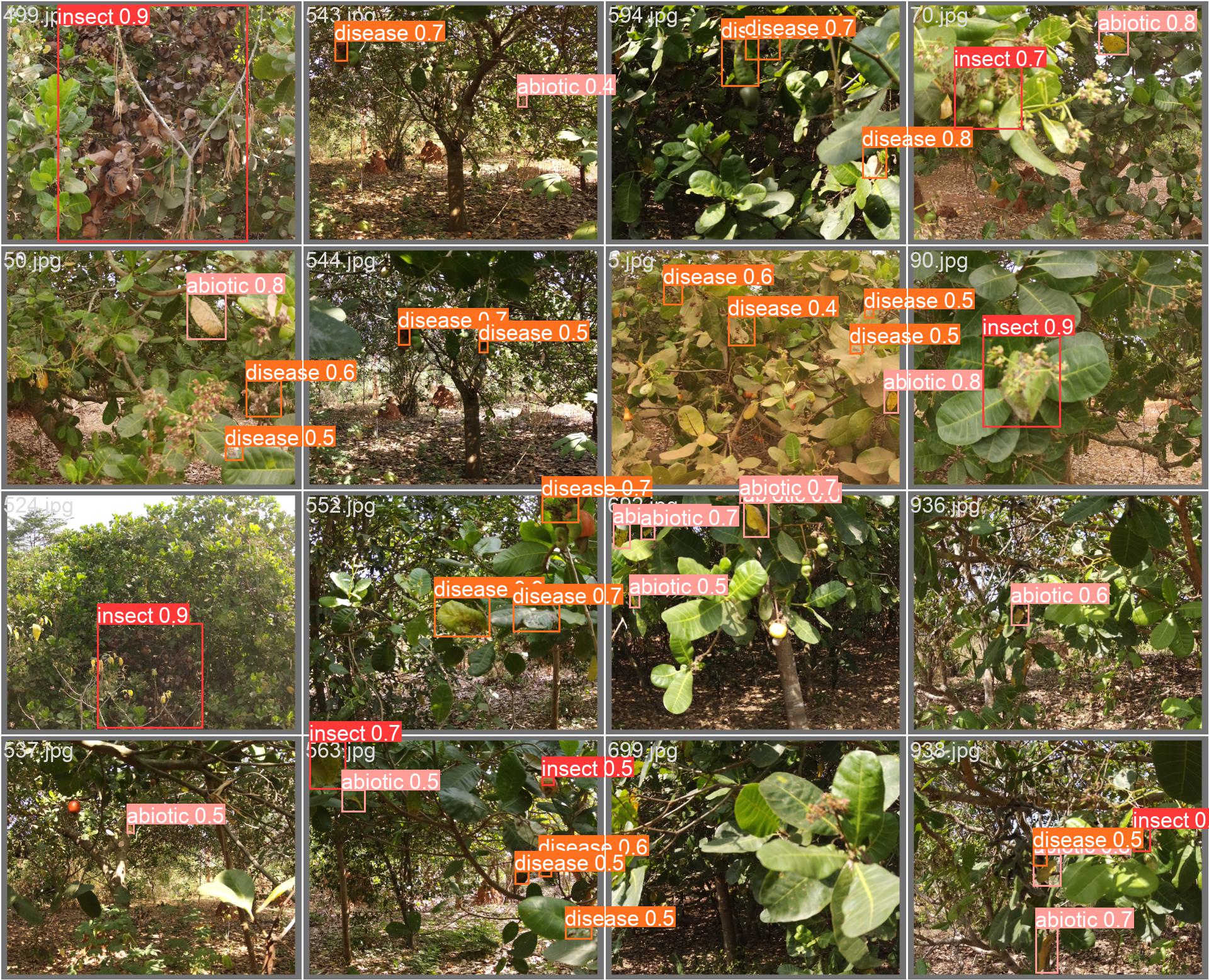}
    \caption{Sample instances from the annotated dataset.}
    \label{fig:pred}
\end{figure}

\clearpage
\section{CADI AI Software}
\begin{figure}
    \centering
    \includegraphics[width=.33\textwidth]{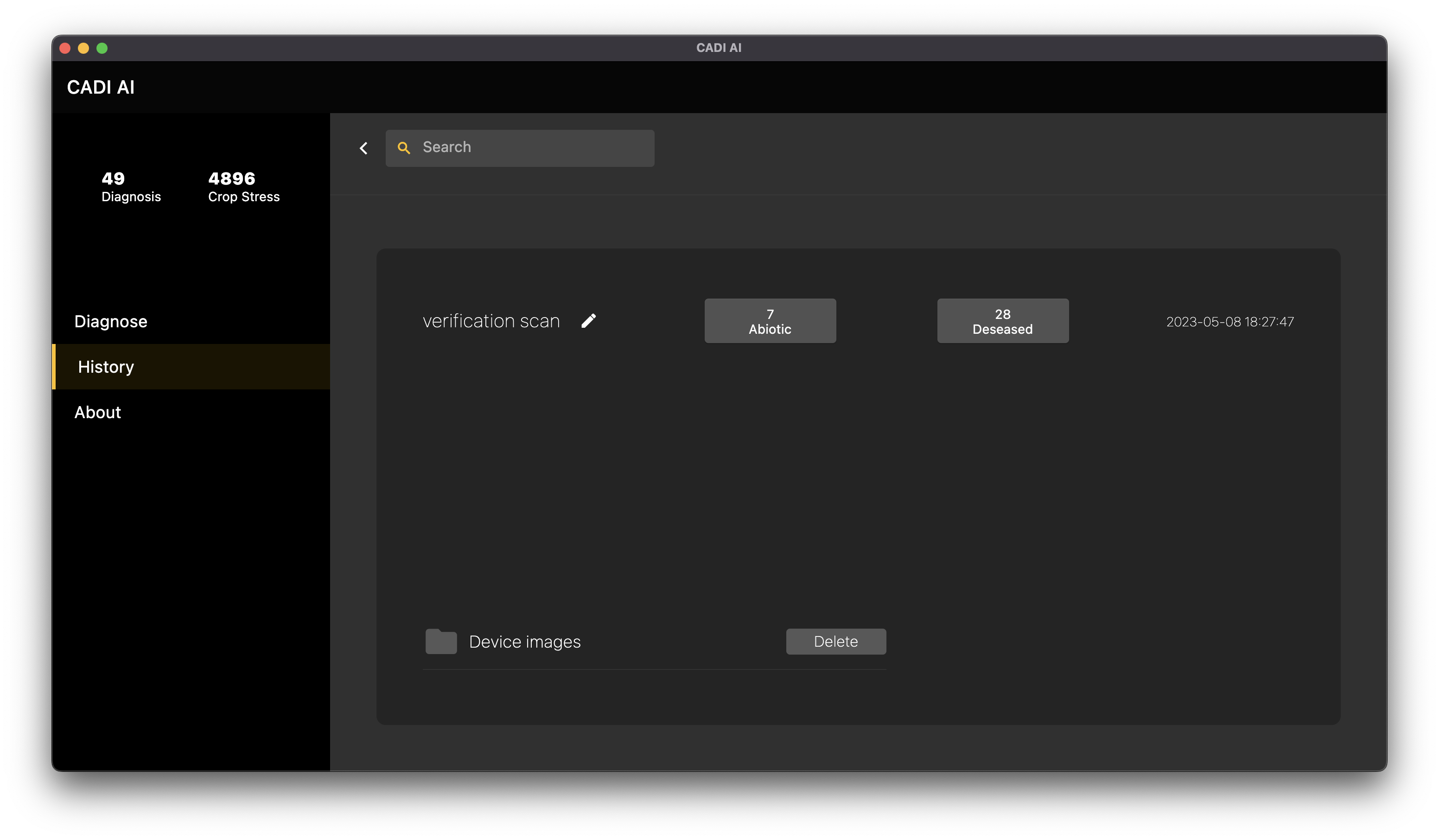}
    \includegraphics[width=.33\textwidth]{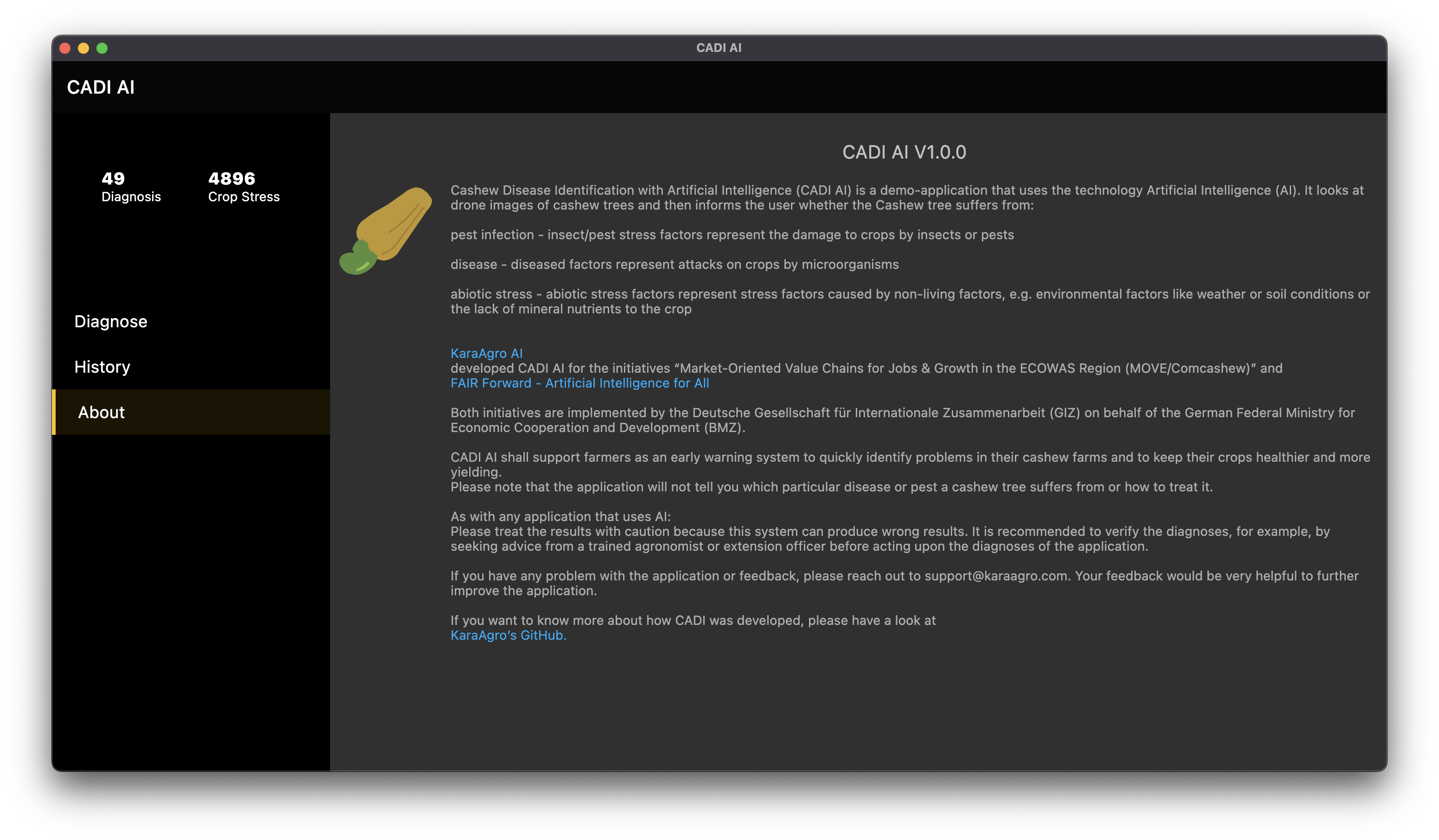}
    \includegraphics[width=.33\textwidth]{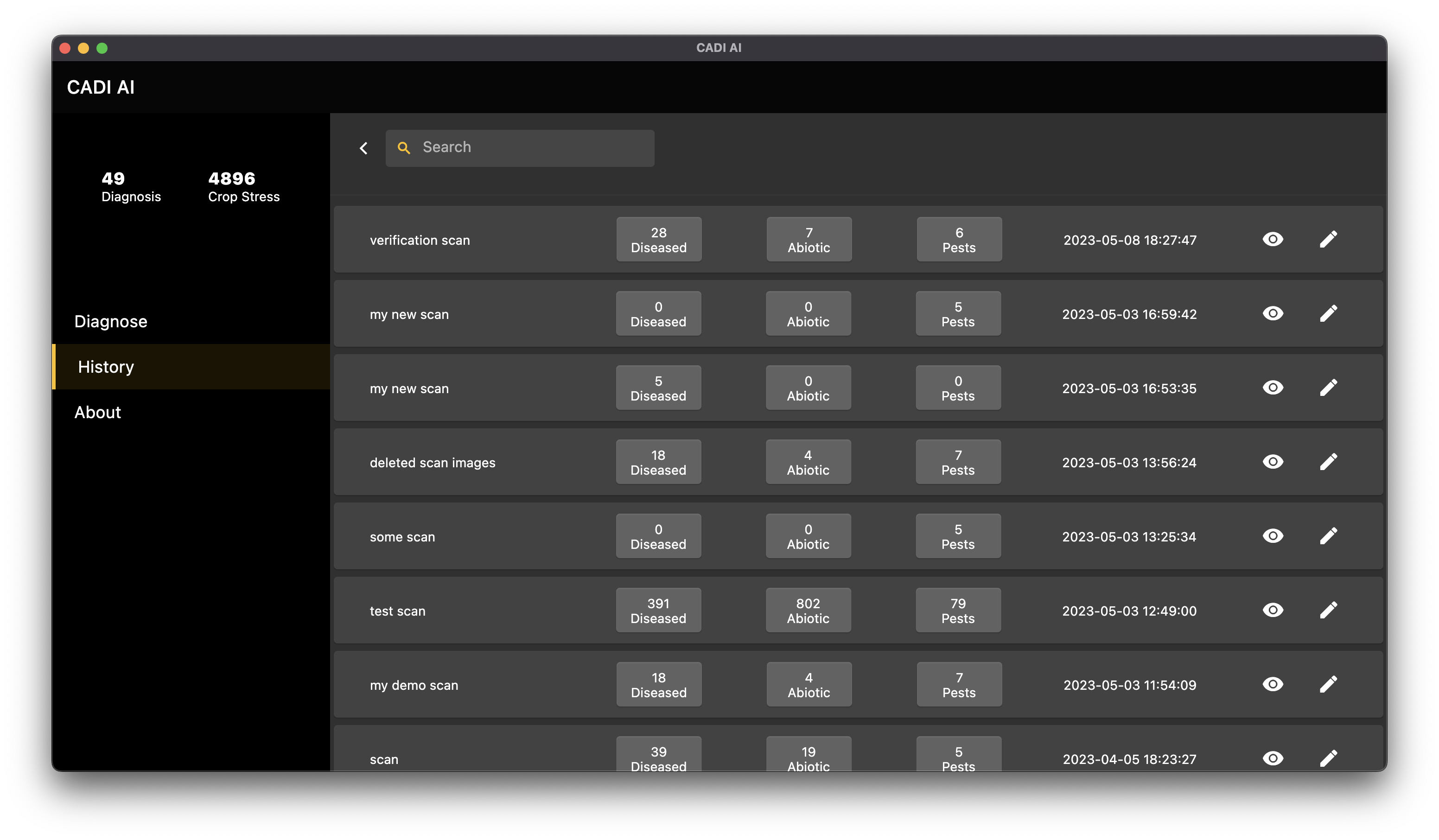}
    \\
    \includegraphics[width=.33\textwidth]{assets/app/stress_locator.png}
    \includegraphics[width=.33\textwidth]{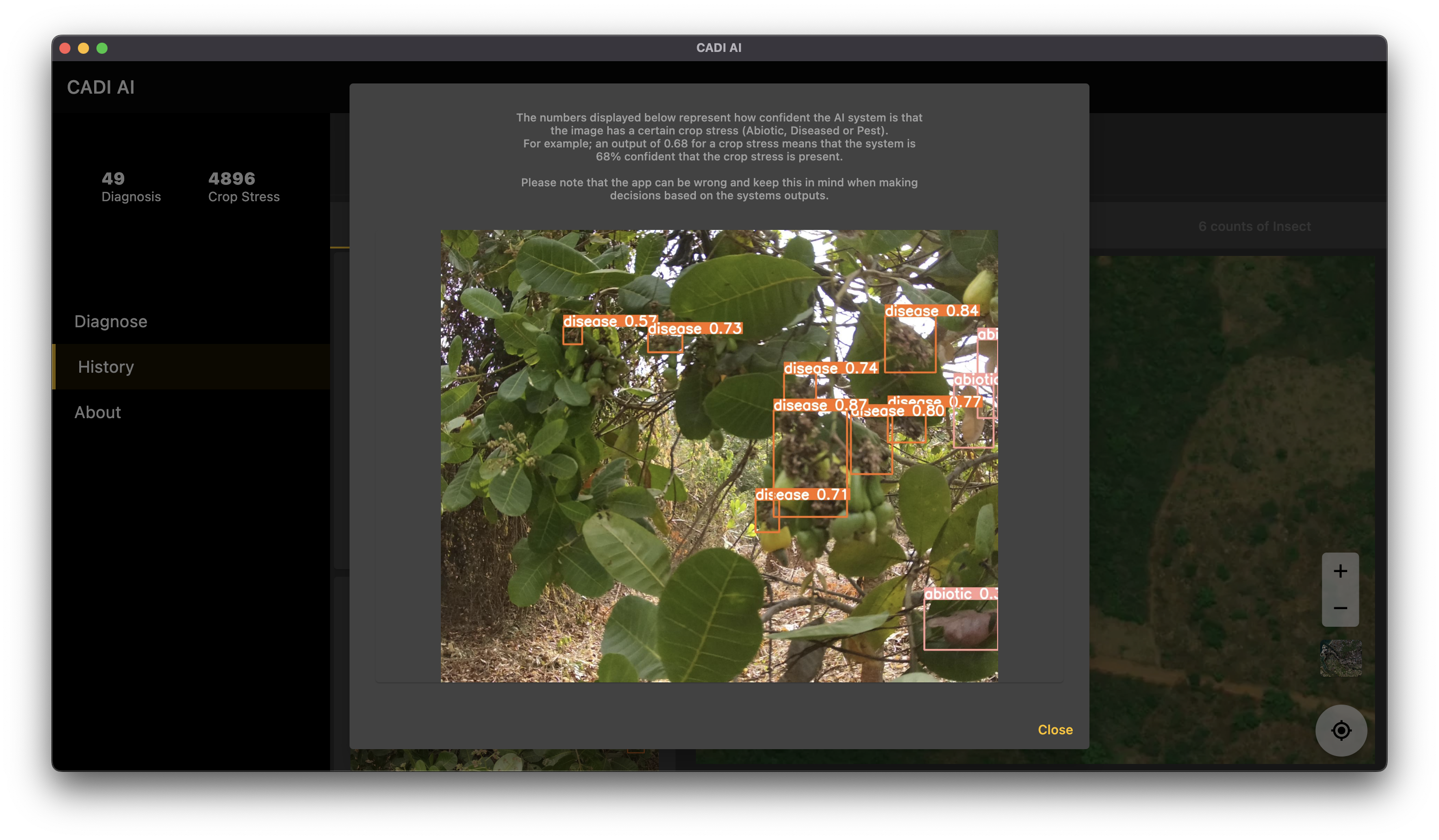}
    \includegraphics[width=.33\textwidth]{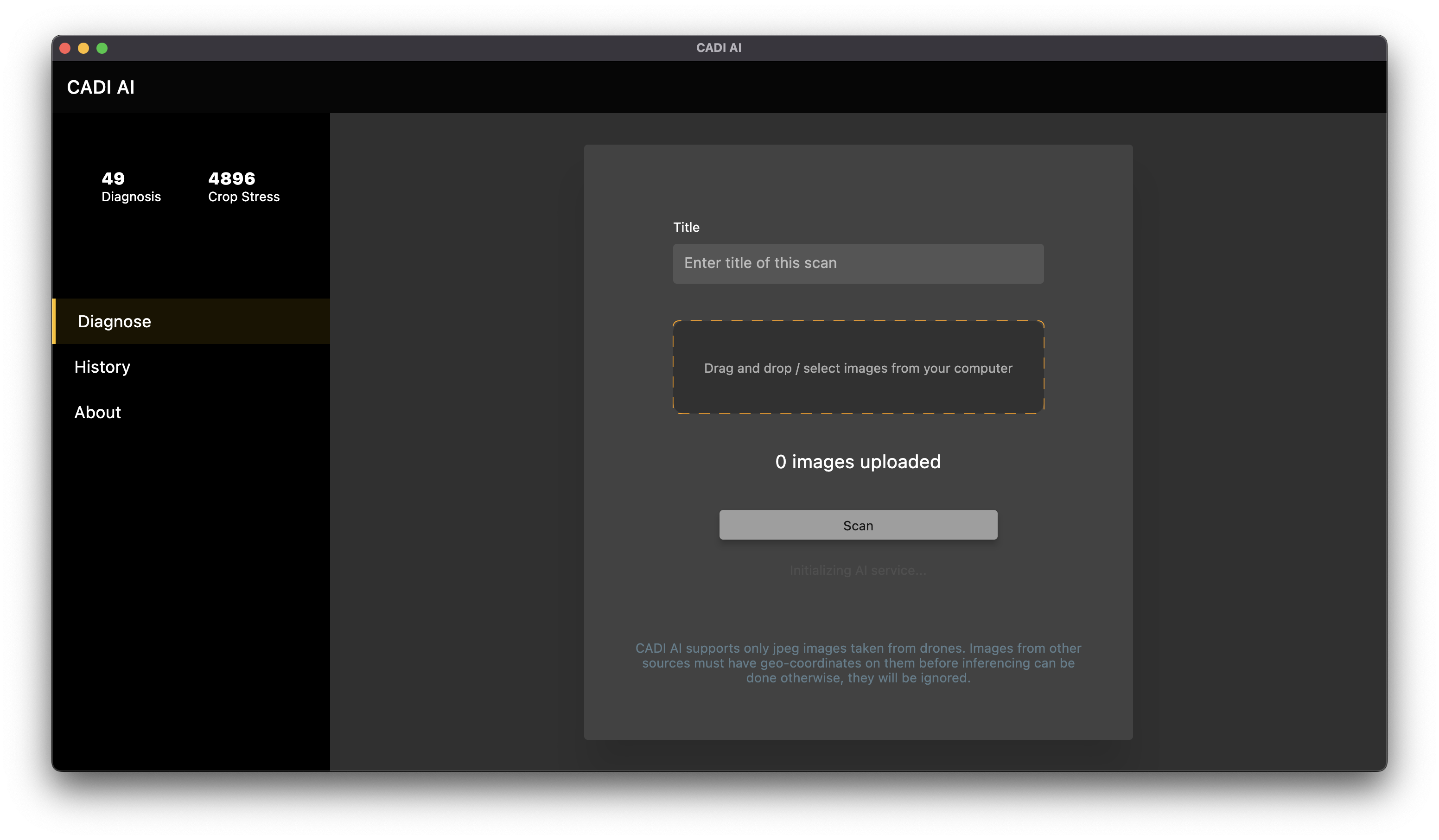}
    \caption{Screenshots of the final app (from left to right, top to bottom): About page, edit window, project history, stress locator, stress analysis, upload screen.}
    \label{fig:app}
\end{figure}
The software is named CADI AI; an abbreviation for Cashew Disease
Identification with Artificial Intelligence. CADI AI has the following
features: 
\begin{itemize}
    \item A desktop interface for accessing all other features of the application
    \item A diagnosis widget where a user could select multiple number of images
for diagnoses without limit
    \item A history widget which shows a history list of all scans made
    \item A widget for viewing all images uploaded in a particular scan with a
satellite map that shows where the images were taken
\end{itemize}

CADI AI does not rely on an internet connection to work. It is, in an ideal
world, a totally offline application. Because the application spins up a
local flask server that enables  communication with the CNN model, there is
sometimes a need to connect to the internet to download metadata
which are then cashed. The local cache can expire after some time and
create the need for a re-sync using an internet connection.

Note that CADI AI only runs scans on images that are having coordinates in
their Exif metadata (Exchangeable Image Format). 

The Exif specifications define a pair of file types, mainly intended for recording technical details associated with digital photography \cite{exif-format}.

Exif is a metadata standard that defines formats for sharing metadata related to images, sound, and ancillary tags used by digital cameras, scanners and other systems that handle image recorded by digital cameras. 

In this case, it is required that images selected for scan in CADI AI must have metadata following the Exif standard, and must have geographical location coordinates in the metadata before these images can be scanned by the CNN model.

This is because the application shows location pins on a satellite map 
that enables farmers to locate the region of the farm where a
disease, abiotic stress or pest is found present. Without GPS coordinates or geographical location coordinates, farmers would be oblivious to where exactly issues are found in their farms.

The development of CADI AI was started in-house at \anon{KaraAgro AI} and the
codebase was later published as an open source repository welcoming
contributions from the open source community and serving up the application
to the general public on github \cite{ai_2023}

\section{Resource Coordinates}
\begin{table}[!h]
    \centering
    \caption{Table Showing URLs to Open Source Resources}
    \begin{tabular}{ll}
    \toprule
        \textbf{} & \textbf{Location} \\ \midrule
        \textbf{Data} & \anon{\url{https://www.kaggle.com/datasets/karaagroaiprojects/cadi-ai}} \\
        & \anon{\url{https://huggingface.co/datasets/KaraAgroAI/CADI-AI}} \\ 
        \textbf{Model} & \anon{\url{https://huggingface.co/KaraAgroAI/CADI-AI}} \\
        \textbf{App Code and Executable} & \anon{\url{https://github.com/karaagro/cadi-ai}} \\ \bottomrule
    \end{tabular}
\end{table}

\section{Contribution and Commercial Use}
As an open-source project under the GNU Affero General Public License(GNU AGPL)\cite{gnu}, the CADI AI project encourages contributions from the wider community. Users are free to modify and build upon the existing resources, tailoring them to their unique needs. Moreover, the resources can be utilized for commercial purposes, providing attribution to the original creators is duly given. This flexibility fosters collaboration and allows the project's benefits to extend to diverse domains and industries.

\section{Datasheet}
\label{sec:datasheet}
\input{datasheet}


\end{document}

%% file: datasheet.tex
\hypertarget{motivation}{%
\subsection{\texorpdfstring{\textbf{3.1
Motivation}}{3.1 Motivation}}\label{motivation}}

\emph{\textbf{For what purpose was the data set created? Was there a
specific task in mind?}}

The creation of this dataset represents a first contribution of drone
data to the field of cashew crop research: Providing an open and
accessible resource of high-quality, well-labeled drone imagery
collected from Ghana Bono-Region, this dataset will offer data
scientists, researchers, and social entrepreneurs within Sub-Saharan
Africa and beyond, opportunities for innovative machine learning
experiments and the development of solutions for \textbf{infield cashew
crop disease diagnosis and spatial analysis}.

\emph{\textbf{Was there a specific gap that needed to be filled? Please
provide a description.}}

Yes. The threat of pests and diseases to the agricultural sector in
Ghana is a constant concern, with climate change contributing to the
potential for new and more damaging types of outbreaks (Yeboah et al.,
2023). Based on multi-stakeholder engagements conducted by \anon{KaraAgro AI},
also with women smallholder cashew farmers, stakeholders have identified
pest and disease detection and yield estimation as critical concerns.

However, the current methods of identifying agricultural pest and
disease outbreaks, such as land surveys and on-site observations by
individuals, are limited in their effectiveness and efficiency. Thus,
there is a need for more innovative and efficient solutions to improve
the monitoring and management of crop health. This highlights a gap in
the available tools and resources, which can be addressed through the
use of advanced technologies such as machine learning and image
analysis.

The creation of an open and accessible cashew dataset with well-labeled,
curated, and prepared imagery can provide a valuable resource for data
scientists, researchers, and social entrepreneurs to develop innovative
solutions towards infield pest and disease detection and yield
estimation.

\textbf{Who created this data set (e.g. which team, research group) and
on behalf of which entity (e.g. company, institution, organization)?}

The dataset was created by a team of data scientists from the \anon{KaraAgro
AI Foundation}, with support from agricultural scientists and officers.

\textbf{Who funded the creation of the data set? If there is an
associated grant, please provide the name of the grantor and the grant
name and number.}

The creation of this dataset was made possible through funding of the
Deutsche Gesellschaft für Internationale Zusammenarbeit (GIZ) through
their projects
``\href{https://www.giz.de/en/worldwide/108524.html}{Market-Oriented
Value Chains for Jobs \& Growth in the ECOWAS Region (MOVE)}'' and
"\href{https://www.bmz-digital.global/en/overview-of-initiatives/fair-forward/}{FAIR
Forward - Artificial Intelligence for All}", which GIZ implements on
behalf the German Federal Ministry for Economic Cooperation and
Development (BMZ).\\
The MOVE initiative aims to support market-oriented and resilient value
chains that contribute to the creation of income and employment in the
ECOWAS region\\
The FAIR Forward initiative aims to promote a more open, inclusive, and
sustainable approach to AI on an international level, by partnering with
countries such as Ghana, India, Indonesia, Kenya, Rwanda, South Africa
and Uganda.

\hypertarget{composition}{%
\subsection{\texorpdfstring{\textbf{3.2
Composition}}{3.2 Composition}}\label{composition}}

\emph{\textbf{What do the instances that comprise the data set represent
( e.g., documents, photos, people, countries)?}}

Each instance in the dataset includes crop image (JPEG), image status
(Disease, Abiotic, and Insect), file type (images and bounding box
annotations) and location (this variable though is without values).

\emph{\textbf{Are there multiple types of instances (e.g., movies,
users, and ratings; people and interactions between them; nodes and
edges)? Please provide a description.}}

No. There is only one instance type, which represents cashew crops based
on drone images with various attributes.

\emph{\textbf{Does the data set contain all possible instances or is it
a sample ( not necessarily random) of instances from a larger set? If
the data set is a sample, then what is the larger set? Is the sample
representative of the larger set (e.g., geographic coverage)? If so,
please describe how this representation was validated/verified. If it is
not representative of the larger set, please describe why not (e.g., to
cover a more diverse range of instances, because instances were withheld
or unavailable).}}

The dataset contains various instances that were captured in the Bono
region, which is renowned for its cashew production. The data was
collected in two rounds: The first data collection happened in November
2022, the second in January 2023.\\
The data captured represents cashew data from a year where cashew
blooming was particularly late. Given that the data was collected
punctually only twice, it might be that not all blooming variations of
Cashews have been captured, potentially influencing the variety of the
collected data.

Thus, the dataset collected is not representative. While more continuous
data collection across various regions during the blooming cycle could
have been beneficial, we still consider our dataset to be sufficiently
diverse. This is due to the inclusion of different maturity stages,
camera angles, time of capture, and various types of stress morphology.

\emph{\textbf{What data does each instance consist of? ``Raw'' data
(e.g., unprocessed text or images) or features? In either case, please
provide a description.}}

Each instance includes: the crop image and location (gps coordinates).

\textbf{\emph{Is there a label or target associated with each instance?
If so, please provide a description}.}

Each instance is associated with a class label based on the status of
the crop. The labels are ``\emph{insect/pest}'', ``\emph{disease}'' and
``\emph{abiotic}'' respectively as depicted in Figure 1:

\begin{itemize}
\item
  \begin{quote}
  \textbf{Insect/ pest} stress factors represent the damage to crops by
  insects or pests
  \end{quote}
\item
  \begin{quote}
  \textbf{Diseased} factors represent attacks on crops by
  microorganisms.
  \end{quote}
\item
  \begin{quote}
  \textbf{Abiotic} stress factors represent stress factors caused by
  non-living factors, e.g. environmental factors like weather or soil
  conditions or the lack of mineral nutrients to the crop.
  \end{quote}
\end{itemize}

The decision to use the labels "abiotic", "disease", and "insect" for
our object detection task was recommended by an agricultural scientist
with expertise in crop health and disease management, \anon{Dr. Torkpor
Stephen} from \anon{University of Ghana}.

It is important to note that while these labels provide a general
categorization of crop damage, they may not fully capture the complexity
of the underlying causes. In addition, the labels may not be exhaustive
and other types of damage may not be captured by these categories. As
with any dataset, users should be aware of the limitations and context
of the labels used and exercise caution when interpreting the results of
models trained on this data.

Examples of the limitations and complexities involved includes

\begin{itemize}
\item
  \begin{quote}
  A plant may exhibit symptoms of both insect damage and disease, making
  it difficult to assign a single label to the damage.
  \end{quote}
\item
  \begin{quote}
  Damage caused by abiotic factors such as drought or nutrient
  deficiency may be similar to damage caused by disease or insect
  infestation, leading to confusion when assigning labels.
  \end{quote}
\item
  \begin{quote}
  Damage caused by multiple factors may not fit neatly into a single
  label category, requiring more nuanced and complex labeling.
  \end{quote}
\item
  \begin{quote}
  Different species of insects or diseases may cause similar damage to
  crops, making it difficult to distinguish between them using only the
  three labels.
  \end{quote}
\item
  \begin{quote}
  Other factors such as environmental stress, mechanical damage, or
  chemical exposure may also cause damage to crops, but may not be
  captured by the current labels.
  \end{quote}
\end{itemize}

After extensive discussions, the project team decided to opt for more
general labels, e.g. ``pest'', instead of specific labels for certain
diseases, e.g. Helopeltis. The reasons for this decision were related to
the AI use case for which the data was planned to be further used (a
more general early-warning system for farmers) and also a weighting of
the available resources, e.g. for collecting and annotating data.

\begin{figure}
    \centering
    \includegraphics[width=1.93229in,height=2.03125in]{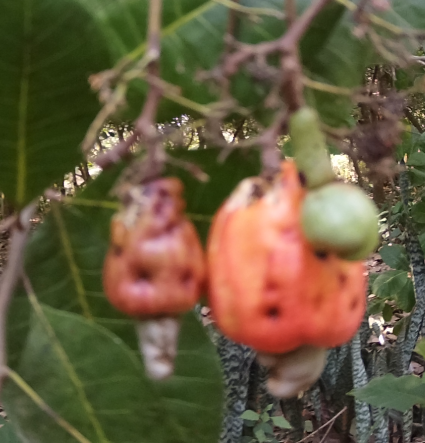}\includegraphics[width=1.93229in,height=2.03125in]{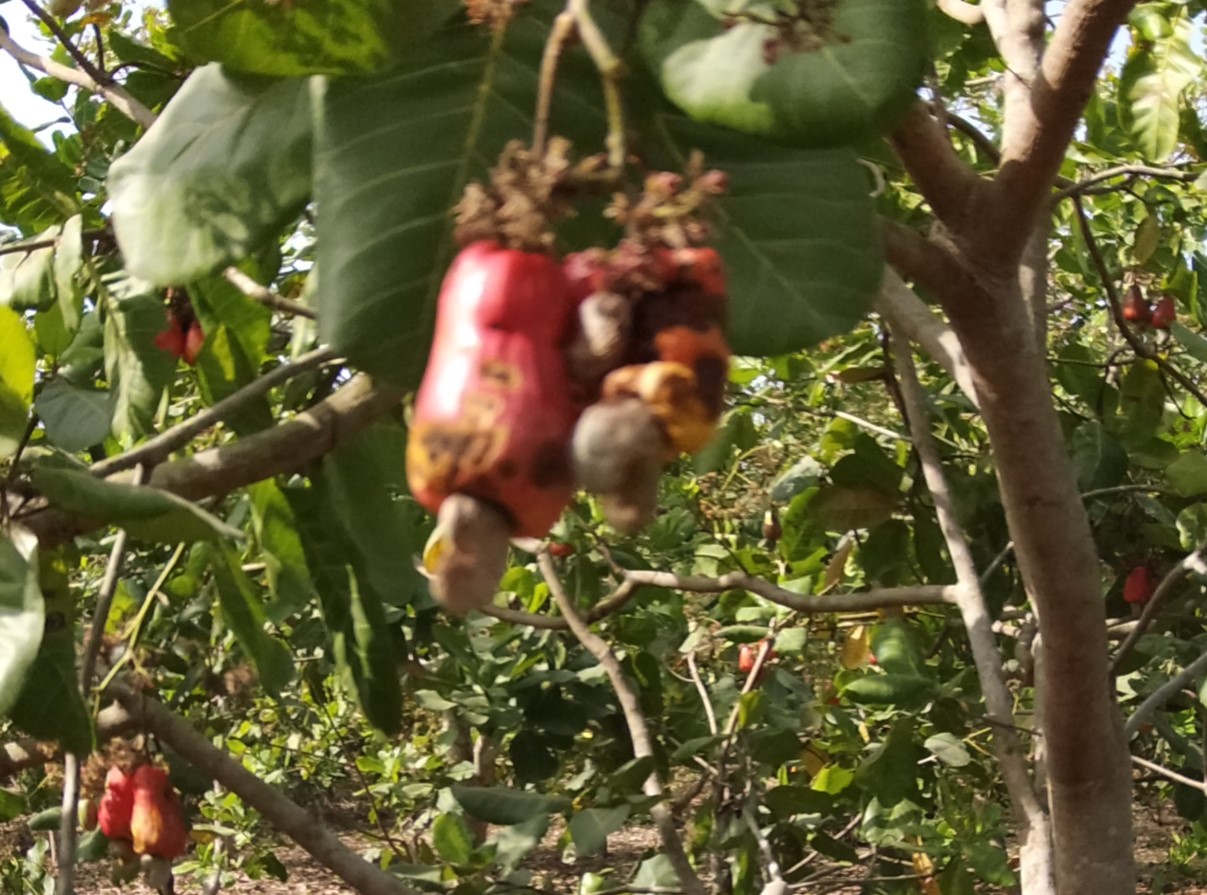}\includegraphics[width=1.93229in,height=2.03125in]{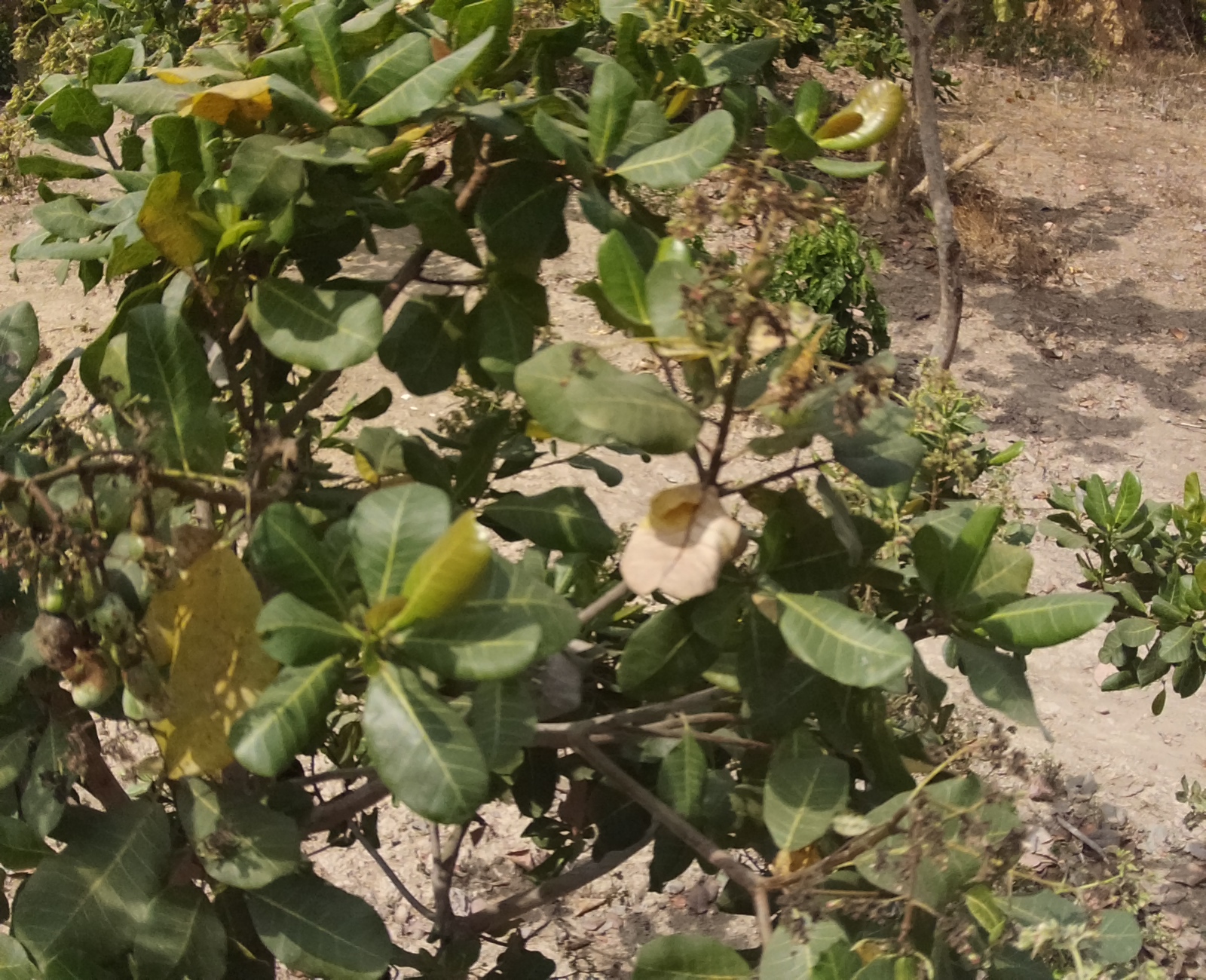}
    \caption{Class labels associated with data (from left to right): Insect, disease, abiotic}
    \label{fig:crops}
\end{figure}

\emph{\textbf{Can you provide a brief description of the number of
images and instances in the dataset, as well as any skewness or
imbalances in the data?}}

The dataset is significantly skewed towards the abiotic class, which
could potentially introduce bias into machine learning models trained on
the data. To address this issue, we took experimental measures during
model development, such as augmenting other classes to balance the data.
However, we found that preserving the skewness was also important, as it
reflected the higher occurrence of abiotic factors in a typical farm.
This approach helped the model recognize the prevalence of abiotic
factors without overemphasizing their importance in predicting other
classes.

\emph{\textbf{Is any information missing from individual instances? If
so, please provide a description, explaining why this information is
missing ( e.g., because it was unavailable). This does not include
intentionally removed information, but might include, e.g., redacted
text.}}

No, the data set contains all the required information.

\emph{\textbf{Is the data set self-contained?}}

Yes, the data is self-contained, it does not rely on any other external
sources.

\emph{\textbf{Please provide descriptions of all external resources and
any restrictions associated with them, as well as links or other access
points, as appropriate. Does the data set contain data that might be
considered confidential (e.g., data that is protected by legal privilege
or by doctor-patient confidentiality, data that includes the content of
individuals' nonpublic communications)? If so, please provide a
description.}}

No, the datasheet is not confidential and it is self-contained.

\emph{\textbf{Does the data set contain data that, if viewed directly,
might be offensive, insulting, threatening, or might otherwise cause
anxiety? If so, please describe why.}}

No.

\emph{\textbf{Does the data set relate to people?}}

No. Given the slight possibility to use the GPS location of cashew trees
to identify individual farms, the project team decided to strip the GPS
location data in the published dataset.

\hypertarget{collection-process}{%
\subsection{\texorpdfstring{\textbf{3.3 Collection
process}}{3.3 Collection process}}\label{collection-process}}

\emph{\textbf{How was the data associated with each instance acquired?
Was the data directly observable (e.g., raw text, movie ratings),
reported by subjects (e.g., survey responses), or indirectly
inferred/derived from other data (e.g., part-of-speech tags, model-based
guesses for age, or language)? If data was reported by subjects or
indirectly inferred/derived from other data, was the data
validated/verified? If so, please describe how.}}

The data associated with each instance was acquired from Bono region
cashew farms in Ghana.

\emph{\textbf{What mechanisms or procedures were used to collect the
data ( e.g., hardware apparatus or sensor, manual human curation,
software program, software API)? How were these mechanisms or procedures
validated?}}

The images for the cashew data collection process were captured using a
drone that was flown manually. The drone was flown at different
altitudes to ensure that comprehensive information about the cashew
crops was gathered. The photos of the cashew crop were taken at
different angles with altitudes ranging from 2 to 10 meters. This
altitude range provides a good balance between capturing a close-up view
of the fruits and their growth stages and a wider perspective that
allows for variation.

\emph{\textbf{What quality management systems were put in place to
ensure the validity, accuracy, and reliability of data collected?}}

\begin{enumerate}
\def\labelenumi{\arabic{enumi}.}
\item
  \begin{quote}
  Careful selection of flight altitude and angle to ensure comprehensive
  data collection.
  \end{quote}
\item
  \begin{quote}
  Use of high-quality drone cameras to capture images of the cashew
  crops.
  \end{quote}
\item
  \begin{quote}
  Careful storage and recording of images, including GPS coordinates and
  timestamps, to provide a record of location and time.
  \end{quote}
\item
  \begin{quote}
  Exclusion of blurry or indistinct images, as well as those with
  defects such as low lighting, to maintain the quality of the dataset.
  \end{quote}
\end{enumerate}

\emph{\textbf{What were the challenges faced in using drones for data
collection and how did they impact the dataset?}}

Challenges faced in data collection using drones:

\begin{itemize}
\item
  \begin{quote}
  Limitations of battery life and flight time
  \end{quote}
\item
  \begin{quote}
  Weather conditions affecting drone performance
  \end{quote}
\item
  \begin{quote}
  Difficulties in navigating terrain with obstacles
  \end{quote}
\item
  \begin{quote}
  Drone whirling up leaves and thus making steady pictures difficult
  \end{quote}
\end{itemize}

Impacts on the dataset:

\begin{itemize}
\item
  \begin{quote}
  Limited coverage area due to battery life and flight time restrictions
  \end{quote}
\item
  \begin{quote}
  Incomplete data due to weather conditions and obstacles
  \end{quote}
\item
  \begin{quote}
  Need for manual inspection and correction of data to ensure accuracy
  \end{quote}
\end{itemize}

To overcome these challenges, the team implemented measures such as
carefully selecting flight paths, checking weather conditions, regularly
calibrating and maintaining drones, and manually inspecting and
correcting data (example, if collected data seemed blurry or not
visible)

Despite these challenges, the use of drones likely provided a more
efficient and cost-effective means of data collection compared to
traditional methods (traditional methods may involve manual inspections
or measurements of crops.) However, the limitations of drone technology
and the need for careful data inspection should also be considered when
analyzing and interpreting the dataset.

\emph{\textbf{Who was involved in the data collection process (e.g.,
students, crowd workers, contractors) and how were they compensated
(e.g., how much were crowd workers paid)?}}

The \anon{KaraAgro AI} team took upon themselves to collect this data.

\emph{\textbf{Over what time frame was the data collected?}}

The data was collected in two rounds: The first data collection happened
in November 2022, the second in January 2023.

\emph{\textbf{Does}} \emph{\textbf{this time frame match the creation
time frame of the data associated with the instances (e.g., a recent
crawl of old news articles)? If not, please describe the time frame in
which the data associated with the instances was created.}}

\emph{\textbf{Were any ethical review processes conducted (e.g., by an
institutional review board)? If so, please describe these review
processes, including the outcomes, as well as a link or other access
point to any supporting documentation.}}

No, there were no ethical review processes conducted.

\emph{\textbf{Does the data set relate to people? If not, you may skip
the remaining questions in this section.}}

No, the data set does not relate to people.

\hypertarget{preprocessing-cleaning-labeling}{%
\subsection{\texorpdfstring{\textbf{3.4 Preprocessing/ Cleaning/
labeling}}{3.4 Preprocessing/ Cleaning/ labeling}}\label{preprocessing-cleaning-labeling}}

\emph{\textbf{Was any preprocessing/cleaning/labeling of the data done
(e.g., discretization or bucketing, tokenization, part-of-speech
tagging, SIFT feature extraction, removal of instances, processing of
missing values)? If so, please provide a description. If not, you may
skip the remainder of the questions in this section.}}

Yes, preprocessing and labeling of the data were done during the data
annotation stage using annotation tools (makesense.ai). Preprocessing of
the data involved removing crop images in which human figures or faces
were accidentally captured. Also blurry images were deleted.

\emph{\textbf{How was it ensured that the annotation of data was
performed accurately and efficiently, and what methods were used to
validate the data and ensure that the annotations were consistent and of
high quality?}}

The data was labeled by data scientists of \anon{KaraAgro AI} who worked on
this project. To ensure the accurate and efficient annotation of data,
the team used advanced annotating tools (makesense.ai, roboflow) that
offered various annotation formats (xml, yolo). Before the annotation
process began, an expert in Agricultural Science reviewed the cashew
images and provided comprehensive training on the annotation process,
including appropriate labels (abiotic, insect, and diseased) to assign
to each image.

During the annotation process, the team was diligent in checking the
images and ensuring that they aligned with the correct labels. Any
inconsistent images were sent to the expert for further analysis and
suggestions. The team only included clear and high-quality images in the
annotated dataset and excluded blurry, indistinct, or defective images
to maintain the quality of the dataset. The team also checked and
removed any images not related to the desired crop to prevent
inaccuracies or confusion in the analysis.

Following the annotation process, the team conducted a thorough peer
review of the annotated cashew images to ensure the quality and accuracy
of the annotations made by team members. This step was crucial in
ensuring that all annotations were consistent and comprehensive, and
that the annotated dataset was of high quality.

\emph{\textbf{Was the ``raw'' data saved in addition to the
preprocessed/cleaned/labeled data (e.g., to support unanticipated future
uses)?}}

The raw unprocessed data (consisting of labeled images) has been saved.

\emph{\textbf{Is the software used to preprocess/clean/label the
instances available? If so, please provide a link or other access
point.}}

Yes, the annotation tool \emph{makesense} can be accessed
\href{https://www.makesense.ai}{here}.

\hypertarget{distribution}{%
\subsection{\texorpdfstring{\textbf{3.5
Distribution}}{3.5 Distribution}}\label{distribution}}

\emph{\textbf{Will the dataset be distributed to third parties outside
of the entity (e.g., company, institution, organization) on behalf of
which the dataset was created? If so, please provide a description.}}

Yes, the dataset will be distributed to third parties outside of the
entity since it will be made publicly available for different companies,
institutions, and organizations.

\emph{\textbf{How will the dataset be distributed (e.g., tarball on the
website, API, GitHub)}}

The dataset can be distributed on Kaggle and Dataverse.

\textbf{\emph{When will the dataset be distributed?}}

Distribution is planned for May 2023.

\emph{\textbf{What license (if any) is it distributed under? Are there
any copyrights on the data?}}

The data will be licensed under the GNU AGPL license where the credit
must be given to the creator, the data can be used for commercial use
and adaptations and it will be shared under identical terms.

\emph{\textbf{Do any export controls or other regulatory restrictions
apply to the dataset or to individual instances? If so, please describe
these restrictions, and provide a link or other access point to, or
otherwise reproduce, any supporting documentation.}}

No, there are no fees or regulatory restrictions that apply to this
dataset.

\hypertarget{uses}{%
\subsection{\texorpdfstring{\textbf{3.6 Uses}}{3.6 Uses}}\label{uses}}

\emph{\textbf{Has the dataset been used for any tasks already? If so,
please provide a description.}}

The dataset was utilized for an object detection task, where a model was
trained to recognize areas affected by abiotic factors, diseases, and
insect infestations of cashew fields.

\emph{\textbf{Is there a repository that links to any or all papers or
systems that use the dataset? If so, please provide a link or other
access point. What (other) tasks could the dataset be used for?}}

\emph{\textbf{What were the main challenges faced by the team during the
data collection and annotation process, and how were these challenges
overcome to ensure the quality of the data collected?}}

During the data collection and annotation process, the team faced
several challenges, such as incomplete or inconsistent labeling,
difficulty in distinguishing between similar classes, and limited
availability of experts for domain-specific knowledge. To address these
challenges, the team implemented several strategies:

\begin{itemize}
\item
  \begin{quote}
  The team established clear guidelines for labeling by inviting
  \anon{Dr.Stephen Torkpor} from \anon{University of Ghana} who is an Agricultural
  Scientist to provide extensive training to team annotators to ensure
  consistency and accuracy.
  \end{quote}
\item
  \begin{quote}
  The team shared ambiguous or uncertain labeling cases with \anon{Dr. Stephen
  Torkpor} of \anon{University of Ghana} for clarification.
  \end{quote}
\item
  \begin{quote}
  The team conducted regular team-internal evaluations of the annotated
  data to ensure that the dataset was of high quality and suitable for
  the intended use.
  \end{quote}
\end{itemize}

\textbf{Below are some guidelines for labeling or annotating the data
set by team.}

\begin{itemize}
\item
  \begin{quote}
  Provide enough space to capture the affected area without cutting any
  part off, but avoid introducing too much space.
  \end{quote}
\item
  \begin{quote}
  If possible, zoom into the affected area to ensure accurate
  annotation.
  \end{quote}
\end{itemize}

Annotation of large affected areas

\begin{itemize}
\item
  \begin{quote}
  Avoid including too much gap between two affected areas. Treat a wide
  gap as another annotation.
  \end{quote}
\item
  \begin{quote}
  To ensure completeness, try to annotate every possible instance as
  long as it is visible.
  \end{quote}
\end{itemize}

Careful annotation

\begin{itemize}
\item
  \begin{quote}
  Even for images that appear to have no affected areas, zoom in and
  carefully examine the image to ensure nothing is missed. It is better
  to have an annotation than to remove the image altogether. However, if
  truly nothing exists, discard the image from the database.
  \end{quote}
\end{itemize}

Avoid human faces

\begin{itemize}
\item
  \begin{quote}
  Avoid using images with human faces whenever possible.
  \end{quote}
\end{itemize}

Cropping out human faces

\begin{itemize}
\item
  \begin{quote}
  If an image with a human face is necessary, but the face is not
  centered, consider cropping it out and replacing the original image
  with the cropped version.
  \end{quote}
\end{itemize}

\emph{\textbf{Is there anything about the composition of the dataset or
the way it was collected and pre-processed/cleaned/labeled that might
impact future uses? For example, is there anything that a future user
might need to know to avoid uses that could result in unfair treatment
of individuals or groups (e.g., stereotyping, quality of service issues)
or other undesirable harms (e.g., financial harms, legal risks) If so,
please provide a description? Is there anything a future user could do
to mitigate these undesirable harms?}}

None that the \anon{KaraAgro AI} team is aware of.

\hypertarget{maintenance}{%
\subsection{\texorpdfstring{\textbf{3.7
Maintenance}}{3.7 Maintenance}}\label{maintenance}}

\emph{\textbf{Who will be supporting/hosting/maintaining the dataset?}}

\anon{KaraAgro AI} will support, host, and maintain the dataset.

\emph{\textbf{How can the dataset owner/curator/manager be contacted
(e.g., email address)?}}

\anon{Darlington Akogo} can be contacted on his email address -
\anon{\href{mailto:darlington@gudra-studio.com}{darlington@gudra-studio.com}}.

\emph{\textbf{Is there an erratum? If so, please provide a link or other
access point.}}

No

\emph{\textbf{Will the dataset be updated (e.g., correcting labeling
errors, adding new instances, deleting instances)? If so, please
describe how often, by whom, and how updates will be communicated to
users (e.g., mailing list, GitHub)?}}

The dataset will be open sourced such that people can contribute to it.
Any changes to the dataset will be documented in a form of dataset
updates

\emph{\textbf{If the dataset relates to people, are there applicable
limits on the retention of the data associated with the instances (e.g.,
were individuals in question told that their data would be retained for
a fixed time and then deleted)? If so, please describe these limits and
explain how they will be enforced.}}

Not applicable.

\emph{\textbf{Will older versions of the dataset continue to be
supported/hosted/maintained? If so, please describe how. If not, please
describe how its obsolescence will be communicated to users.}}

The dataset will be hosted on public servers. Older version links will
still be available in a form of versioning.

\textbf{\emph{If others want to extend/augment/build on/contribute to
the dataset, is there a mechanism for them to do so? If so, please
provide a description. Will these contributions be validated/verified?
If so, please describe how. If not, why not? Is there a process for
communicating/distributing these contributions to other users? If so,
please provide a description. Other researchers are allowed to extend
this dataset.}}

The interested researchers should send an email to
\anon{darlington@gudra-studio.com owned by Darlington Akogo} and they will be
able to discuss the dataset building, extensions, and contributions.

\emph{{[} Refer to sustainability strategy documentation on how to
contribute to this project {]}}

\textbf{References}

\begin{enumerate}
\def\labelenumi{\arabic{enumi}.}
\item
  \begin{quote}
  Patrick Ateah Yeboah, Bismarck Yelfogle Guba, Emmanuel K. Derbile,
  Smallholder cashew production and household livelihoods in the
  transition zone of Ghana, Geo: Geography and Environment,
  10.1002/geo2.120, 10, 1, (2023).
  \end{quote}
\end{enumerate}